\documentclass[10pt,twocolumn,letterpaper]{article}

\usepackage[pagenumbers]{cvpr}

\usepackage{amsmath,amssymb,amsfonts}
\usepackage{graphicx}
\usepackage{booktabs}
\usepackage{multirow}
\usepackage{hyperref}
\usepackage{xcolor}
\usepackage{subcaption}
\usepackage{pgfplots}
\pgfplotsset{compat=1.18}
\usepackage{tikz}
\usetikzlibrary{calc}
\usepackage{enumitem}
\usepackage[numbers,sort&compress]{natbib}
\newcommand{\etal}{\textit{et al.}}

\title{\LARGE \bf Do Open-Loop Metrics Predict Closed-Loop Driving? \\
A Cross-Benchmark Correlation Study of NAVSIM and Bench2Drive}

\author{
    Yiru Wang,
    Anqing Jiang,
    Shuo Wang, 
    Yuwen Heng, 
    Hai Yang, 
    Yang Chen, 
    Hao Sun
    \thanks{
    This work was supported by Bosch Corporate Research. (\textit{Corresponding author: hao.sun4@cn.bosch.com})} 
    \thanks{Yiru Wang,
            Anqing Jiang,
            Shuo Wang,
            Yuwen Heng,
            Hai Yang, 
            Yang Chen, 
            Hao Sun are with Bosch Corporate Research, Bosch (China) Investment Ltd., Shanghai, China. }
}

\date{}

\begin{document}

\maketitle

\begin{figure*}[t]
\centering
\begin{tikzpicture}
\begin{axis}[
  name=plotA,
  width=0.33\textwidth,
  height=0.25\textwidth,
  xlabel={\small Open-Loop L2 $\downarrow$},
  ylabel={\small Bench2Drive DS $\uparrow$},
  xmin=0.6, xmax=1.8,
  ymin=15, ymax=75,
  grid=major,
  grid style={gray!15},
  mark size=2.5pt,
  title={\small (a) L2 vs.\ DS: $\rho{=}{-}0.36$},
  title style={yshift=-2pt},
]
\addplot[only marks, mark=*, red!50!black, mark size=2.5pt] coordinates {
  (1.70, 23.63) (0.73, 37.72) (0.91, 39.42)
  (0.95, 39.88) (1.01, 42.91) (0.82, 52.80) (0.92, 65.89)
};
\end{axis}

\begin{axis}[
  name=plotB,
  at={($(plotA.east)+(1.2cm,0)$)},
  anchor=west,
  width=0.33\textwidth,
  height=0.25\textwidth,
  xlabel={\small NAVSIM PDMS $\uparrow$},
  ylabel={},
  xmin=82, xmax=95,
  ymin=40, ymax=95,
  grid=major,
  grid style={gray!15},
  mark size=2.5pt,
  title={\small (b) PDMS vs.\ DS: $\rho{=}0.90$},
  title style={yshift=-2pt},
]
\addplot[only marks, mark=*, blue!70!black, mark size=2.5pt] coordinates {
  (83.4, 45.81) (86.5, 59.95) (88.3, 61.71)
  (88.6, 73.86) (89.3, 76.15) (91.6, 66.77)
  (92.0, 89.15) (93.5, 83.02)
};
\addplot[dashed, blue!40!black, domain=82:95] {-155.5 + 2.59*x};
\end{axis}

\begin{axis}[
  name=plotC,
  at={($(plotB.east)+(1.2cm,0)$)},
  anchor=west,
  width=0.33\textwidth,
  height=0.25\textwidth,
  ybar,
  bar width=10pt,
  enlarge x limits=0.15,
  xlabel={\small NAVSIM Sub-Metric},
  ylabel={\small Spearman $\rho$ with DS},
  ymin=-0.1, ymax=1.15,
  xtick=data,
  xticklabels={NC, TTC, DAC, EP, PDMS},
  xticklabel style={font=\scriptsize},
  ymajorgrids=true,
  xmajorgrids=false,
  grid style={gray!15},
  title={\small (c) Which Sub-Metric Predicts DS?},
  title style={yshift=-2pt},
  nodes near coords,
  nodes near coords style={font=\tiny, anchor=south, /pgf/number format/fixed},
  every node near coord/.append style={yshift=1pt},
  clip=false,
]
\addplot[fill=gray!40, draw=gray!60] coordinates {
  (1, 0.45) (2, 0.59) (3, 0.71) (4, 0.83) (5, 0.90)
};
\draw[fill=blue!50, draw=blue!70!black]
  ([xshift=-5pt]axis cs:5,0) rectangle ([xshift=5pt]axis cs:5,0.90);
\draw[fill=teal!60, draw=teal!80!black]
  ([xshift=-5pt]axis cs:4,0) rectangle ([xshift=5pt]axis cs:4,0.83);
\end{axis}
\end{tikzpicture}
\caption{\textbf{Can open-loop metrics predict closed-loop driving?} Each point represents one method evaluated on both benchmarks; $n$ denotes the number of such paired methods. (a)~Traditional L2 displacement shows no correlation with closed-loop Driving Score ($n{=}7$ paired methods, CARLA Leaderboard~v2 protocol). (b)~NAVSIM's safety-aware PDMS shows strong positive correlation ($n{=}8$ paired methods), but with ranking inversions. (c)~\textbf{Not all sub-metrics are equal}: among individual NAVSIM components, Ego Progress (EP) is the strongest single predictor of closed-loop DS ($\rho{=}0.83$), substantially exceeding the collision metric NC ($\rho{=}0.45$). The aggregate PDMS slightly exceeds EP alone ($\rho{=}0.90$), but as we show, the same correlation can be achieved by a much simpler 3-metric formula (\S\ref{sec:regression}).}
\label{fig:teaser}
\end{figure*}
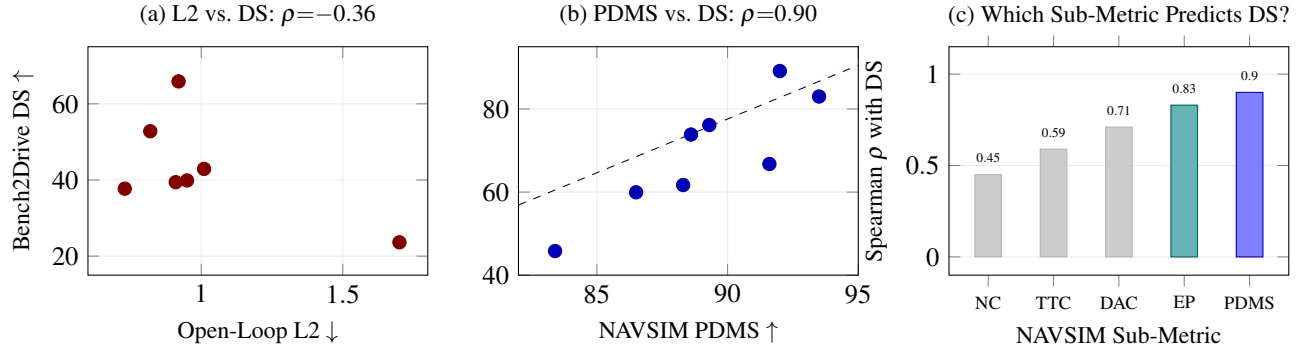

\begin{abstract}
Open-loop evaluation offers fast, reproducible assessment of autonomous driving planners, but its ability to predict real closed-loop driving performance remains questionable. Prior work has shown that traditional open-loop metrics such as Average Displacement Error (ADE) and Final Displacement Error (FDE) exhibit \emph{no reliable correlation} with closed-loop Driving Score. In this paper, we ask whether the more recent, safety-aware open-loop metrics introduced by NAVSIM~v2 can bridge this gap. By systematically cross-referencing published results from 15 state-of-the-art methods across NAVSIM (open-loop) and Bench2Drive (closed-loop), we compile a paired dataset of open-loop sub-metrics and closed-loop performance, yielding 8 methods with complete paired data. Our analysis reveals three key findings: (1)~the aggregate NAVSIM PDM Score shows a strong positive but \emph{non-monotonic} correlation with Bench2Drive Driving Score (Spearman~$\rho = 0.90$, $p = 0.002$), with clear ranking inversions; (2)~among individual NAVSIM sub-metrics, Ego Progress (EP) is the strongest single predictor of closed-loop success ($\rho = 0.83$), substantially exceeding the safety-critical collision metric NC ($\rho = 0.45$); (3)~the safety--progress trade-off manifests differently in open-loop and closed-loop: methods that maximize safety at the expense of progress rank highly in NAVSIM but underperform in closed-loop due to timeout and slow-driving penalties. We further demonstrate that a much simpler 3-metric formula (CL-Proxy~$=$~NC$\times$DAC$\times$EP) matches the predictive power of the full 5-metric PDMS at the same Spearman $\rho{=}0.90$ on our paired sample of $n{=}8$ methods, suggesting that within current state-of-the-art methods---where TTC and Comfort approach saturation---these two sub-metrics add little marginal information for closed-loop ranking. Additionally, we identify the \emph{snowball effect}---where small open-loop deviations compound into closed-loop failures---as a candidate mechanism for the residual gap.

\end{abstract}

\section{Introduction}
\label{sec:intro}

Evaluating autonomous driving planners in closed-loop simulation---where the ego vehicle executes its planned trajectory step by step and the environment reacts accordingly---is the gold standard for assessing driving competence. However, closed-loop evaluation incurs substantial computational cost (hours per model on CARLA~\citep{dosovitskiy2017carla}), suffers from high variance across runs, and requires complex simulation infrastructure. Open-loop evaluation, which assesses a single predicted trajectory against ground truth, offers orders-of-magnitude faster turnaround and near-perfect reproducibility.

The fundamental question is: \textbf{Can open-loop metrics reliably predict closed-loop driving performance?} As illustrated in Figure~\ref{fig:teaser}, the answer depends critically on \emph{which} open-loop metric is used.

Prior work answers this question negatively for traditional metrics. Dauner~\etal~\cite{dauner2023parting} demonstrated that ADE and FDE---the dominant open-loop metrics for trajectory prediction---show \emph{no meaningful correlation} with closed-loop driving quality. In the nuPlan Challenge 2023, the top-ranked open-loop method performed poorly in closed-loop evaluation. Li~\etal~\cite{li2024ego} further exposed the fragility of traditional open-loop metrics by showing that a simple MLP using only ego status (speed and acceleration) can achieve state-of-the-art open-loop performance on nuScenes, without any scene understanding.

Recognizing these shortcomings, the NAVSIM benchmark~\citep{dauner2024navsim} introduced a fundamentally different open-loop evaluation paradigm. Instead of measuring trajectory displacement, NAVSIM employs a suite of \emph{safety-aware, rule-based sub-metrics}---including collision detection (NC), drivable area compliance (DAC), time-to-collision (TTC), ego progress (EP), and comfort metrics---aggregated via a multiplicative penalty structure that mimics the ``one crash ends the run'' semantics of closed-loop evaluation. The extended version, NAVSIM~v2, further adds driving direction compliance (DDC), traffic light compliance (TLC), lane keeping (LK), and extended comfort metrics, resulting in the Extended PDM Score (EPDMS).

Meanwhile, the Bench2Drive benchmark~\citep{jia2024bench2drive} provides a standardized closed-loop evaluation protocol on CARLA~v2, measuring Driving Score (DS) and Success Rate (SR) across 220 diverse routes with 44 interactive scenarios. As the community has matured, an increasing number of methods now report results on \emph{both} NAVSIM and Bench2Drive, creating an unprecedented opportunity for cross-benchmark correlation analysis.

\textbf{Contributions.} In this paper, we:
\begin{enumerate}[leftmargin=*]
  \item Compile a \textbf{cross-benchmark paired dataset} by systematically cross-referencing published results from 15 methods that report on both NAVSIM and Bench2Drive, yielding 8 complete paired data points (\S\ref{sec:data}).
  \item Show that \textbf{NAVSIM's safety-aware aggregate score (PDMS) exhibits positive but non-monotonic correlation with closed-loop DS}, with clear ranking inversions caused by the safety--progress trade-off (\S\ref{sec:pdms_analysis}).
  \item Identify \textbf{Ego Progress (EP) as the strongest individual sub-metric predictor} of closed-loop success, and reveal that the safety--progress trade-off is the primary source of ranking disagreement between open-loop and closed-loop evaluations (\S\ref{sec:submetric}).
  \item Propose a \textbf{simplified closed-loop proxy} (CL-Proxy) using only 3 of the 5 PDMS sub-metrics that matches the rank correlation of the full PDMS on our paired data, suggesting that TTC and Comfort add little marginal predictive value among current top methods (\S\ref{sec:regression}).

\end{enumerate}

\section{Related Work}
\label{sec:related}

\paragraph{Open-Loop Evaluation.}
Open-loop evaluation has been the default paradigm in trajectory prediction and motion planning. Traditional metrics such as ADE, FDE, and miss rate measure geometric deviation between predicted and ground-truth trajectories~\citep{caesar2020nuscenes,wilson2023argoverse}. However, these metrics do not account for safety, rule compliance, or physical feasibility. The nuPlan benchmark~\citep{karnchanachari2024nuplan} introduced more holistic metrics but maintained an open-loop protocol. NAVSIM~\citep{dauner2024navsim} represented a paradigm shift by introducing non-reactive simulation with safety-aware scoring, designed to approximate closed-loop semantics within an open-loop framework. NAVSIM~v2 extended this with additional rule-compliance and comfort metrics. Despite these advances, no systematic study has quantified how well NAVSIM metrics predict actual closed-loop performance.

\paragraph{Closed-Loop Evaluation.}
Closed-loop benchmarks evaluate driving agents in interactive simulation. The CARLA Leaderboard~\citep{dosovitskiy2017carla} was among the first, but suffered from reproducibility issues. Bench2Drive~\citep{jia2024bench2drive} addressed this by providing standardized short routes with diverse interactive scenarios, measuring Driving Score as the product of route completion and infraction penalties. The recently proposed Bench2Drive-VL~\citep{jia2026bench2drivevl} extends this to vision-language models. Despite their fidelity, closed-loop benchmarks remain expensive and high-variance, motivating research into open-loop proxies.

\paragraph{Open-Loop vs.\ Closed-Loop Gap.}
Dauner~\etal~\cite{dauner2023parting} provided the first systematic analysis showing that open-loop trajectory metrics do not predict closed-loop performance, attributing the gap to compounding errors, non-reactive agents, and distribution shift. Li~\etal~\cite{li2024ego} further demonstrated that open-loop metrics can be trivially gamed. More recently, the Waymo team~\cite{waymo2025scaling} showed that \emph{within a single architecture family}, open-loop loss and closed-loop failure rates follow a power-law relationship when scaling data and compute---but this relationship breaks down across different architectures. Our work bridges these findings by asking whether NAVSIM's \emph{multi-dimensional, safety-aware} metrics can serve as a more reliable open-loop proxy across diverse architectures.

\paragraph{End-to-End Autonomous Driving.}
Recent end-to-end planners have increasingly been evaluated on both open-loop and closed-loop benchmarks, enabling the cross-benchmark analysis we conduct. Notable examples include UniAD~\citep{hu2023uniad}, Hydra-MDP~\citep{li2024hydramdp}, DiffusionDrive~\citep{liao2025diffusiondrive}, Hydra-NeXt~\citep{li2025hydranext}, GoalFlow~\citep{xing2025goalflow}, SparseDriveV2~\citep{sun2026sparsedrivev2}, SafeDrive~\citep{kim2026safedrive}, DriveTransformer~\citep{jia2025drivetransformer}, VADv2~\citep{chen2024vadv2}, WoTE~\citep{li2025wote}, and DriveSuprim~\citep{sima2025drivesuprim}.

\section{Background}
\label{sec:background}

\subsection{NAVSIM Open-Loop Metrics}
\label{sec:navsim_metrics}

NAVSIM evaluates predicted trajectories using a set of sub-metrics, aggregated into the PDM Score (PDMS) and the Extended PDM Score (EPDMS).

\paragraph{PDMS.} The PDM Score combines penalty metrics (multiplicative gates) with weighted average metrics:
\begin{equation}
  \text{PDMS} = \underbrace{\text{NC} \times \text{DAC}}_{\text{penalty gates}} \times \frac{5 \cdot \text{TTC} + 2 \cdot \text{C} + 5 \cdot \text{EP}}{12},
  \label{eq:pdms}
\end{equation}
where NC (No at-fault Collision) and DAC (Drivable Area Compliance) act as binary/ternary gates that zero out the score upon violation, while TTC (Time-to-Collision), C (Comfort), and EP (Ego Progress) contribute to a weighted average.

\paragraph{EPDMS.} The Extended PDM Score adds four sub-metrics and modifies the aggregation:
\begin{equation}
  \text{EPDMS} = \prod_{m \in \mathcal{P}} m \;\cdot\; \frac{\textstyle\sum_{m \in \mathcal{W}} w_m \cdot m}{\textstyle\sum_{m \in \mathcal{W}} w_m},
  \label{eq:epdms}
\end{equation}
where $\mathcal{P} = \{\text{NC}, \text{DAC}, \text{DDC}, \text{TLC}\}$ are penalty gates, and $\mathcal{W}$ contains EP\,(5), TTC\,(5), LK\,(2), HC\,(2), EC\,(2) with weights in parentheses. EPDMS also applies human-behavior filtering: if the human driver also violates a rule, the predicted trajectory is not penalized.

\subsection{Bench2Drive Closed-Loop Metrics}
\label{sec:b2d_metrics}

Bench2Drive~\citep{jia2024bench2drive} evaluates closed-loop performance on 220 short routes in CARLA across 44 interactive scenarios.

\paragraph{Driving Score (DS).} The primary metric combines route completion with infraction penalties:
\begin{equation}
  \text{DS} = \frac{1}{n} \sum_{i=1}^{n} \text{RC}_i \cdot \prod_{j=1}^{m_i} p_j^{(i)},
  \label{eq:ds}
\end{equation}
where $\text{RC}_i$ is the route completion percentage and $p_j^{(i)}$ is the penalty factor for the $j$-th infraction (e.g., 0.50 for pedestrian collision, 0.60 for vehicle collision, 0.70 for running red lights, 0.70 for driving too slowly).

\paragraph{Success Rate (SR).} The fraction of routes completed without any infractions:
\begin{equation}
  \text{SR} = \frac{n_{\text{success}}}{n_{\text{total}}}.
  \label{eq:sr}
\end{equation}

\paragraph{Evaluation Protocols.}
Bench2Drive supports two evaluation protocols that differ in how slow-driving infractions are handled~\citep{li2025hydranext}. Under the \emph{CARLA~v2 protocol}, minimum-speed infractions are accumulated multiplicatively into the Driving Score ($p = 0.70$ each), making DS sensitive to conservative driving. The \emph{Bench2Drive protocol} instead separates speed infractions into a standalone \emph{Efficiency} metric and relaxes the route time limit, yielding systematically higher DS for the same model. For example, Hydra-NeXt reports DS~65.89 under CARLA~v2 vs.\ 73.86 under the Bench2Drive protocol. Unless otherwise noted, we use the Bench2Drive protocol for cross-benchmark correlation analysis (Tables~\ref{tab:pdms_vs_ds}--\ref{tab:submetrics}), as it is the protocol adopted by the majority of recent publications.

\subsection{Structural Similarities and Differences}
\label{sec:structural}

Both NAVSIM and Bench2Drive employ multiplicative penalty structures for safety violations. However, key differences exist:

\begin{itemize}[leftmargin=*]
  \item \textbf{Reactivity}: NAVSIM replays recorded agent trajectories regardless of the ego's actions; Bench2Drive simulates reactive agents.
  \item \textbf{Compounding errors}: NAVSIM evaluates a single 4-second trajectory; Bench2Drive executes trajectories step-by-step over entire routes, accumulating errors.
  \item \textbf{Slow-driving penalty}: Bench2Drive penalizes overly slow driving ($p = 0.70$); NAVSIM has no equivalent penalty---EP measures progress but does not penalize conservatism as harshly.
  \item \textbf{Control conversion}: Bench2Drive requires converting trajectories to vehicle control commands (steering, throttle, brake), introducing an additional error source absent in NAVSIM.
\end{itemize}

These structural differences create systematic biases in the open-to-closed-loop mapping, which we analyze in \S\ref{sec:analysis}.

\section{Cross-Benchmark Paired Dataset}
\label{sec:data}

\begin{table}[ht]
\centering
\caption{Traditional open-loop L2 error vs.\ closed-loop metrics. Data from Hydra-NeXt~\citep{li2025hydranext} Table~1, evaluated under the CARLA Leaderboard~v2 protocol (DS values are systematically lower than the standard Bench2Drive protocol used in Tables~\ref{tab:pdms_vs_ds}--\ref{tab:submetrics}). * denotes methods using expert feature distillation.}

\label{tab:l2_vs_ds}
\small
\begin{tabular}{lccc}
\toprule
Method & Open-loop L2 $\downarrow$ & DS $\uparrow$ & SR $\uparrow$ \\
\midrule
TCP~\cite{wu2022tcp}              & 1.70 & 23.63 & 7.72\%  \\
UniAD~\cite{hu2023uniad}          & 0.73 & 37.72 & 9.54\%  \\
VAD~\cite{jiang2023vad}            & 0.91 & 39.42 & 10.00\% \\
ThinkTwice*~\cite{jia2023thinktwice} & 0.95 & 39.88 & 28.14\% \\
DriveAdapter*~\cite{jia2023driveadapter} & 1.01 & 42.91 & 30.71\% \\
Hydra-MDP~\cite{li2024hydramdp}   & 0.82 & 52.80 & 30.73\% \\
Hydra-NeXt~\cite{li2025hydranext} & 0.92 & 65.89 & 48.20\% \\
\bottomrule
\end{tabular}
\end{table}

\subsection{Data Collection Methodology}

We compile paired open-loop and closed-loop results by systematically cross-referencing published tables from recent papers. Specifically, we extract:
\begin{itemize}[leftmargin=*]
  \item NAVSIM PDMS and sub-metrics from Tables in SparseDriveV2~\citep{sun2026sparsedrivev2}, SafeDrive~\citep{kim2026safedrive}, Hydra-NeXt~\citep{li2025hydranext}, and WoTE~\citep{li2025wote}.
  \item Bench2Drive DS and SR from the same papers, as well as from VADv2~\citep{chen2024vadv2} and DriveSuprim~\citep{sima2025drivesuprim}.
  \item Traditional open-loop L2 error from Hydra-NeXt~\citep{li2025hydranext} Table~\ref{tab:l2_vs_ds}.
\end{itemize}

\begin{table*}[ht]
\centering
\caption{Cross-benchmark results from SparseDriveV2~\citep{sun2026sparsedrivev2}, SafeDrive~\citep{kim2026safedrive}, VADv2~\citep{chen2024vadv2}, and DriveSuprim~\citep{sima2025drivesuprim}. All Bench2Drive results use the standard Bench2Drive evaluation protocol. ``---'' indicates the metric was not reported in the source.}
\label{tab:pdms_vs_ds}
\setlength{\tabcolsep}{5pt}
\small
\begin{tabular}{lccccl}
\toprule
Method & NAVSIM PDMS $\uparrow$ & NAVSIM EPDMS $\uparrow$ & B2D DS $\uparrow$ & B2D SR $\uparrow$ & Source \\
\midrule
UniAD~\cite{hu2023uniad}            & 83.4 & ---  & 45.81 & 16.36\% & SafeDrive T1+T3 \\
VAD~\cite{jiang2023vad}              & ---  & ---  & 42.35 & 15.00\% & SafeDrive T3 \\
BridgeAD~\cite{zhang2025bridgead}      & ---  & ---  & 50.06 & 22.73\% & SafeDrive T3 \\
Hydra-MDP~\cite{li2024hydramdp}      & 86.5 & ---  & 59.95 & 29.82\% & SafeDrive T1+T3 \\
WoTE~\cite{li2025wote}               & 88.3 & ---  & 61.71 & 31.36\% & WoTE T1+T2 \\
DriveDPO~\cite{shang2025drivedpo}    & ---  & ---  & 62.02 & 30.62\% & SafeDrive T3 \\
DriveTransformer~\cite{jia2025drivetransformer} & ---  & ---  & 63.46 & 35.01\% & SparseDriveV2 T4 \\
SafeDrive~\cite{kim2026safedrive}    & 91.6 & 87.5 & 66.77 & 42.40\% & SafeDrive T1+T3 \\
Hydra-NeXt~\cite{li2025hydranext}    & 88.6 & ---  & 73.86 & 50.00\% & Hydra-NeXt T2+T4 \\
VADv2~\cite{chen2024vadv2}           & 89.3 & 85.8 & 76.15 & 50.46\% & VADv2 T2+T5 \\
DriveSuprim~\cite{sima2025drivesuprim} & 93.5 & 87.1 & 83.02 & 60.00\% & DriveSuprim T2+T4 \\
SimLingo~\cite{renz2025simlingo}   & ---  & ---  & 86.02 & 67.27\% & SparseDriveV2 T4 \\
HiP-AD~\cite{tang2025hipad}         & ---  & ---  & 86.77 & 69.09\% & SparseDriveV2 T4 \\
DiffRefiner~\cite{yin2025diffrefiner} & ---  & 87.4 & 87.10 & 71.40\% & Abstract (AAAI'26) \\
SparseDriveV2~\cite{sun2026sparsedrivev2} & 92.0 & 90.1 & \textbf{89.15} & \textbf{70.00\%} & SparseDriveV2 T2+T4 \\
\bottomrule
\end{tabular}
\end{table*}

We match methods that appear in both open-loop and closed-loop tables of the \emph{same paper} or across papers that cite the same original checkpoint/configuration. We exclude methods where the open-loop and closed-loop evaluations used different model configurations or training data.

To ensure the validity of our sub-metric correlation analysis (Table~\ref{tab:submetrics}), we adopt the following configuration-matching policy. For methods reported in third-party survey tables (UniAD, Hydra-MDP, WoTE, Hydra-NeXt, SafeDrive, SparseDriveV2), we use the cross-benchmark numbers as compiled by SparseDriveV2~\citep{sun2026sparsedrivev2}, SafeDrive~\citep{kim2026safedrive}, and Hydra-NeXt~\citep{li2025hydranext}, where consistent model configurations across NAVSIM and Bench2Drive are documented. For VADv2 and DriveSuprim, we extract NAVSIM sub-metrics from each method's own paper (VADv2 Table~2, DriveSuprim Table~2 ViT-L row) and pair them with the Bench2Drive DS reported in the same paper. We acknowledge that single-paper sourcing does not eliminate all configuration drift across benchmarks (e.g., backbone or training-data differences), and we discuss this confounder in \S\ref{sec:discussion}.

\subsection{Paired Data Tables}

\paragraph{Traditional Open-Loop Metrics vs.\ Closed-Loop.}
Table~\ref{tab:l2_vs_ds} presents the traditional L2 displacement error alongside Bench2Drive DS and SR for 7 methods, sourced from Hydra-NeXt~\citep{li2025hydranext}.

\paragraph{NAVSIM PDMS vs.\ Closed-Loop.}
Table~\ref{tab:pdms_vs_ds} presents the full cross-benchmark dataset for methods with both NAVSIM PDMS and Bench2Drive results. Table~\ref{tab:pdms_paired} extracts the subset with complete PDMS and DS data.

\paragraph{NAVSIM Sub-Metrics.}
Table~\ref{tab:submetrics} presents the NAVSIM PDMS sub-metric breakdown alongside Bench2Drive DS for methods where both are available.

\begin{table}[ht]
\centering
\caption{Paired NAVSIM PDMS and Bench2Drive DS ($n=8$). Ordered by PDMS. $\dagger$From the method's own paper.}
\label{tab:pdms_paired}
\small
\begin{tabular}{lccc}
\toprule
Method & PDMS $\uparrow$ & DS $\uparrow$ & SR $\uparrow$ \\
\midrule
UniAD~\cite{hu2023uniad}                 & 83.4 & 45.81 & 16.36\% \\
Hydra-MDP~\cite{li2024hydramdp}          & 86.5 & 59.95 & 29.82\% \\
WoTE~\cite{li2025wote}                   & 88.3 & 61.71 & 31.36\% \\
Hydra-NeXt~\cite{li2025hydranext}        & 88.6 & 73.86 & 50.00\% \\
VADv2$^\dagger$~\cite{chen2024vadv2}     & 89.3 & 76.15 & 50.46\% \\
SafeDrive~\cite{kim2026safedrive}        & 91.6 & 66.77 & 42.40\% \\
SparseDriveV2~\cite{sun2026sparsedrivev2} & 92.0 & 89.15 & 70.00\% \\
DriveSuprim$^\dagger$~\cite{sima2025drivesuprim} & 93.5 & 83.02 & 60.00\% \\
\bottomrule
\end{tabular}
\end{table}

\begin{table}[ht]
\centering
\caption{NAVSIM sub-metrics and Bench2Drive DS. Comfort (C) omitted (all $\geq$99.9). Data sourced from the respective original publications. All 8 methods confirm consistent model configurations across both benchmarks.}
\label{tab:submetrics}
\setlength{\tabcolsep}{3pt}
\small
\begin{tabular}{lcccc|c|cc}
\toprule
\multirow{2}{*}{Method} & \multicolumn{4}{c|}{NAVSIM Sub-Metrics} & \multirow{2}{*}{PDMS} & \multicolumn{2}{c}{Bench2Drive} \\
\cmidrule(lr){2-5} \cmidrule(lr){7-8}
 & NC & DAC & TTC & EP &  & DS & SR \\
\midrule
UniAD~\cite{hu2023uniad}                 & 97.8 & 91.9 & 92.9 & 78.8 & 83.4 & 45.8 & 16.4\% \\
Hydra-MDP~\cite{li2024hydramdp}          & 98.3 & 96.0 & 94.6 & 78.7 & 86.5 & 60.0 & 29.8\% \\
WoTE~\cite{li2025wote}                   & 98.5 & 96.8 & 94.9 & 81.9 & 88.3 & 61.7 & 31.4\% \\
Hydra-NeXt~\cite{li2025hydranext}        & 98.1 & 97.7 & 94.6 & 81.8 & 88.6 & 73.9 & 50.0\% \\
SafeDrive~\cite{kim2026safedrive} & \textbf{99.5} & \textbf{99.0} & \textbf{97.2} & 84.3 & 91.6 & 66.8 & 42.4\% \\
VADv2~\cite{chen2024vadv2} & 98.3 & 97.4 & 95.7 & 82.3 & 89.3 & 76.2 & 50.5\% \\
DriveSuprim~\cite{sima2025drivesuprim} & 98.6 & 98.6 & 95.5 & \textbf{91.3} & \textbf{93.5} & 83.0 & 60.0\% \\
SparseDriveV2~\cite{sun2026sparsedrivev2} & 98.5 & 98.4 & 95.0 & 88.6 & 92.0 & \textbf{89.2} & \textbf{70.0\%} \\
\bottomrule
\end{tabular}
\end{table}

\section{Analysis}
\label{sec:analysis}

\subsection{Traditional Metrics: ADE/FDE Still Do Not Predict Closed-Loop}
\label{sec:trad}

We first confirm, with a broader set of recent methods, the finding of~\cite{dauner2023parting}: traditional displacement-based open-loop metrics do not predict closed-loop performance.

From Table~\ref{tab:l2_vs_ds} (CARLA~v2 protocol; see \S\ref{sec:b2d_metrics} for the difference from the Bench2Drive protocol), UniAD achieves the \emph{lowest} L2 error (0.73) among non-distillation methods, yet its DS (37.72) is the second-worst. Conversely, Hydra-NeXt has a higher L2 (0.92) but achieves the best DS (65.89). Computing the Spearman rank correlation between L2 and DS yields $\rho = -0.36$ ($p = 0.43$), indicating \emph{no significant correlation}. This confirms that traditional open-loop metrics remain unreliable proxies for closed-loop performance, even for newer architectures.

\subsection{NAVSIM PDMS: Positive but Non-Monotonic Correlation}
\label{sec:pdms_analysis}

In contrast to L2, the NAVSIM PDMS shows a substantially stronger relationship with Bench2Drive DS. From the eight paired data points in Table~\ref{tab:pdms_paired}, we compute:
\begin{itemize}[leftmargin=*]
  \item Spearman rank correlation: $\rho = 0.90$ ($p = 0.002$)
  \item Kendall's $\tau$: $\tau = 0.79$ ($p = 0.005$)
\end{itemize}

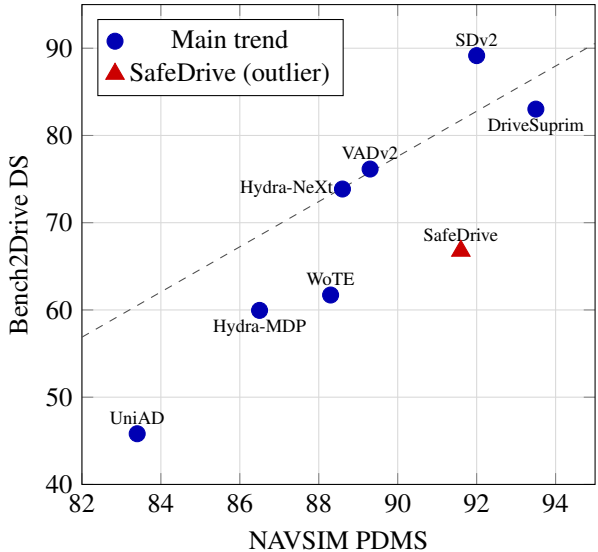
\begin{figure}[ht]
\centering
\begin{tikzpicture}
\begin{axis}[
  width=0.95\columnwidth,
  height=0.90\columnwidth,
  xlabel={NAVSIM PDMS},
  ylabel={Bench2Drive DS},
  xmin=82, xmax=95,
  ymin=40, ymax=95,
  grid=major,
  grid style={gray!30},
  every node near coord/.append style={font=\scriptsize, anchor=south west},
  mark size=2.5pt,
  legend pos=north west,
]
\addplot[only marks, mark=*, blue!70!black, mark size=3pt] coordinates {
  (83.4, 45.81)
  (86.5, 59.95)
  (88.3, 61.71)
  (88.6, 73.86)
  (89.3, 76.15)
  (92.0, 89.15)
  (93.5, 83.02)
};
\addplot[only marks, mark=triangle*, red!80!black, mark size=4pt] coordinates {
  (91.6, 66.77)
};
\node[font=\scriptsize, anchor=south] at (axis cs:83.4,45.81) {UniAD};
\node[font=\scriptsize, anchor=north] at (axis cs:86.5,59.95) {Hydra-MDP};
\node[font=\scriptsize, anchor=south] at (axis cs:88.3,61.71) {WoTE};
\node[font=\scriptsize, anchor=east] at (axis cs:88.6,73.86) {Hydra-NeXt};
\node[font=\scriptsize, anchor=south] at (axis cs:89.3,76.15) {VADv2};
\node[font=\scriptsize, anchor=south] at (axis cs:91.6,66.77) {SafeDrive};
\node[font=\scriptsize, anchor=south] at (axis cs:92.0,89.15) {SDv2};
\node[font=\scriptsize, anchor=north] at (axis cs:93.5,83.02) {DriveSuprim};
\addplot[dashed, gray!60!black, domain=82:95] {-155.5 + 2.59*x};
\legend{Main trend, SafeDrive (outlier)}
\end{axis}
\end{tikzpicture}
\caption{NAVSIM PDMS vs.\ Bench2Drive Driving Score ($n=8$, Spearman $\rho=0.90$, $p=0.002$). SafeDrive (\textcolor{red!80!black}{$\blacktriangle$}) is the primary outlier: it ranks 3rd in PDMS but drops to 5th in DS, due to the safety--progress trade-off analyzed in \S\ref{sec:submetric}.}
\label{fig:scatter_pdms}
\end{figure}

These values indicate a \emph{highly significant positive correlation}, marking a substantial improvement over traditional metrics. However, the correlation is \emph{not perfectly monotonic}: the PDMS ranking and DS ranking disagree on specific methods, as visualized in Figure~\ref{fig:scatter_pdms}.

\paragraph{Ranking Inversion: SafeDrive.}
The most prominent ranking inversion involves SafeDrive. Among the 8~methods, SafeDrive ranks 3rd by PDMS (91.6) but only 5th by DS (66.77)---a two-position drop from open-loop to closed-loop ranking. Similarly, DriveSuprim ranks 1st in PDMS (93.5) but 2nd in DS (83.02), behind SparseDriveV2's DS of 89.15. Conversely, Hydra-NeXt ranks 5th in PDMS (88.6) but 4th in DS (73.86), gaining a position in the closed-loop ranking. We investigate the cause of these inversions next.

\subsection{Sub-Metric Analysis: The Safety--Progress Trade-Off}
\label{sec:submetric}

To understand the SafeDrive ranking inversion, we examine NAVSIM sub-metrics (Table~\ref{tab:submetrics}). SafeDrive achieves \emph{the highest scores across all safety metrics}: NC~99.5, DAC~99.0, TTC~97.2---substantially outperforming all other methods. However, its EP (Ego Progress) is only 84.3, lower than SparseDriveV2's 88.6.

The mechanism is clear: SafeDrive employs fine-grained safety reasoning (PwNC and TwDAC modules) that aggressively avoids collisions and lane departures. This conservative strategy inflates safety sub-metrics in NAVSIM, boosting the aggregate PDMS. However, in Bench2Drive's closed-loop evaluation, this conservatism triggers ``Too Slow'' penalties ($p = 0.70$ per occurrence) and timeouts ($p = 0.70$), significantly reducing the Driving Score.

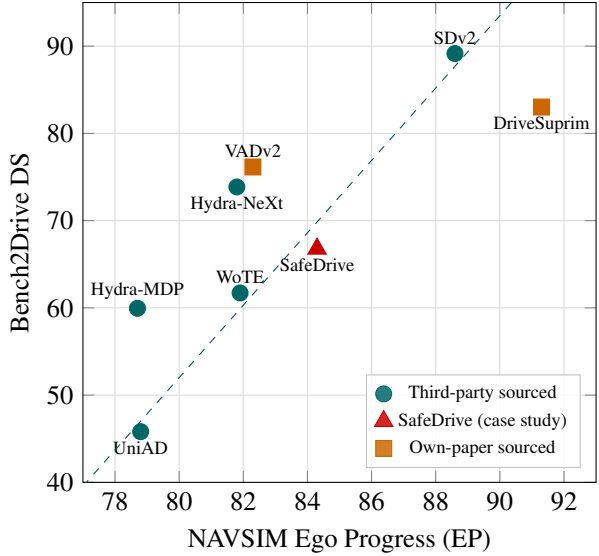
\begin{figure}[ht]
\centering
\begin{tikzpicture}
\begin{axis}[
  width=0.95\columnwidth,
  height=0.9\columnwidth,
  xlabel={NAVSIM Ego Progress (EP)},
  ylabel={Bench2Drive DS},
  xmin=77, xmax=93,
  ymin=40, ymax=95,
  grid=major,
  grid style={gray!30},
  mark size=2.5pt,
  legend pos=south east,
  legend style={font=\scriptsize, fill=white, fill opacity=0.85, draw=gray!50, text opacity=1},
]
\addplot[only marks, mark=*, teal!80!black, mark size=3pt] coordinates {
(78.8, 45.81) (78.7, 59.95) (81.9, 61.71) (81.8, 73.86) (88.6, 89.15)
};

\addplot[only marks, mark=triangle*, red!80!black, mark size=4pt] coordinates {
(84.3, 66.77)
};

\addplot[only marks, mark=square*, orange!80!black, mark size=3pt] coordinates {
(82.3, 76.15) (91.3, 83.02)
};

\node[font=\scriptsize, anchor=north] at (axis cs:78.8,45.81) {UniAD};
\node[font=\scriptsize, anchor=south] at (axis cs:78.7,59.95) {Hydra-MDP};
\node[font=\scriptsize, anchor=south] at (axis cs:81.9,61.71) {WoTE};
\node[font=\scriptsize, anchor=north] at (axis cs:81.8,73.86) {Hydra-NeXt};
\node[font=\scriptsize, anchor=north] at (axis cs:84.3,66.77) {SafeDrive};
\node[font=\scriptsize, anchor=south] at (axis cs:82.3,76.15) {VADv2};
\node[font=\scriptsize, anchor=north] at (axis cs:91.3,83.02) {DriveSuprim};
\node[font=\scriptsize, anchor=south] at (axis cs:88.6,89.15) {SDv2};

\addplot[dashed, teal!60!black, domain=77:93, forget plot] {-280 + 4.15*x};
\legend{Third-party sourced, SafeDrive (case study), Own-paper sourced}
\end{axis}
\end{tikzpicture}
\caption{Ego Progress (EP) vs.\ Bench2Drive Driving Score for the $n{=}8$ methods with complete sub-metric data (Spearman $\rho=0.83$). All methods confirm consistent model configurations across both benchmarks. EP exhibits the strongest correlation among individual sub-metrics.}
\label{fig:scatter_ep}
\end{figure}

\paragraph{EP as the Strongest Individual Predictor.}
We compute the Spearman correlation ($\rho$) between each NAVSIM sub-metric and Bench2Drive DS across the eight methods in Table~\ref{tab:submetrics}: \textbf{EP}~$\rho = 0.83$, DAC~$\rho = 0.71$, TTC~$\rho = 0.59$, NC~$\rho = 0.45$. Figure~\ref{fig:scatter_ep} visualizes the EP--DS relationship.

EP exhibits the highest correlation among individual sub-metrics, although it is slightly below the aggregate PDMS ($\rho=0.90$, Figure~\ref{fig:teaser}c). Remarkably, NC---the most safety-critical metric---shows only a weak-to-moderate correlation ($\rho = 0.45$) with closed-loop DS, well below EP. While extreme safety optimization helps (e.g., DriveSuprim achieves NC=98.6 and SafeDrive NC=99.5), marginal NC gains beyond 98.0 suffer from a ceiling effect and do not guarantee closed-loop success as consistently as Ego Progress does. The fact that EP alone (one sub-metric) achieves $\rho=0.83$ while PDMS (five sub-metrics) achieves $\rho=0.90$ raises a natural question: \emph{which of the five sub-metrics actually contribute predictive value?} We address this in \S\ref{sec:regression}.

\paragraph{Interpretation.}
The safety--progress trade-off operates asymmetrically across the two evaluation paradigms:
\begin{itemize}[leftmargin=*]
  \item \textbf{In NAVSIM (open-loop)}: Safety metrics (NC, DAC, TTC) act as multiplicative gates. Maximizing them dramatically boosts PDMS because even small improvements (e.g., NC from 98.3 to 99.5) eliminate penalty cases. EP, despite being a weighted-average term, has limited leverage on the aggregate score.
  \item \textbf{In Bench2Drive (closed-loop)}: Safety is a \emph{necessary condition} (collisions are penalized), but \emph{route completion} is the dominant factor in DS. Overly conservative driving incurs explicit penalties (Too Slow: 0.70, Timeout: 0.70, Agent Blocked: route termination). EP, which directly reflects the vehicle's willingness to make progress, becomes the critical differentiator.
\end{itemize}

This asymmetry explains why methods optimized for NAVSIM safety metrics (like SafeDrive) can rank highly in open-loop yet underperform in closed-loop.

\section{Toward Better Open-Loop Prediction of Closed-Loop Performance}
\label{sec:regression}

\subsection{A Simpler Equivalent Form: CL-Proxy}

The aggregate PDMS achieves a strong rank correlation with closed-loop DS ($\rho{=}0.90$), but its formula (Eq.~\ref{eq:pdms}) combines five sub-metrics through a multi-stage gate-and-weighted-average pipeline. We ask: \emph{is all this machinery necessary?} Concretely, we investigate whether a simpler aggregation of fewer sub-metrics can recover the same predictive power.

\paragraph{Motivation.}
Our sub-metric correlation analysis (\S\ref{sec:submetric}) revealed two facts that question the design of PDMS:
\begin{enumerate}[leftmargin=*]
  \item TTC shows only moderate correlation with DS ($\rho{=}0.59$), yet is given the same weight as EP in the PDMS weighted average ($5{:}5$).
  \item Comfort (C) is saturated ($\geq{}99.9$\%) across all eight methods and thus contributes essentially zero discriminative information to the ranking.
\end{enumerate}
If TTC and C are non-predictive, removing them should not degrade the rank correlation. We test this hypothesis with a deliberately minimal formula.

\paragraph{Proposed CL-Proxy Score.}
We define the \emph{Closed-Loop Proxy Score} (CL-Proxy) as the simplest multiplicative combination of the three remaining sub-metrics:
\begin{equation}
  \text{CL-Proxy} = \text{NC} \times \text{DAC} \times \text{EP}.
  \label{eq:clproxy}
\end{equation}
This preserves PDMS's multiplicative gate structure (NC and DAC penalize collisions and lane departures), but replaces the weighted average $(5\cdot\text{TTC}{+}2\cdot\text{C}{+}5\cdot\text{EP})/12$ with EP alone. The formula uses 3~sub-metrics instead of 5, has no tunable weights, and no normalization constant. We note the rank correlation is highly robust to functional choice: alternative simplifications such as $\text{DAC}\times\text{EP}$, $(\text{DAC}\cdot\text{EP})^{1/2}$, and even the linear $\text{DAC}+\text{EP}$ all yield identical $\rho{=}0.90$ on our $n{=}8$ data.

\paragraph{Evaluation.}
Table~\ref{tab:loocv} compares the ranking accuracy of PDMS, EP alone, and CL-Proxy against actual Bench2Drive DS rankings for the 8 paired methods.

\begin{table}[h]
\centering
\caption{Ranking prediction accuracy of different open-loop predictors against Bench2Drive DS ($n{=}8$). CL-Proxy uses 3 of PDMS's 5 sub-metrics yet matches the full aggregate. ``\#sm'' = number of sub-metrics; ``Inv.'' = pairwise rank inversions.}
\label{tab:loocv}
\small
\setlength{\tabcolsep}{4pt}
\begin{tabular}{lccc}
\toprule
Predictor & \#sm & Spearman $\rho$ $\uparrow$ & Inv. $\downarrow$ \\
\midrule
PDMS (aggregate)         & 5 & \textbf{0.90} & \textbf{3} \\
EP only                  & 1 & 0.83          & 5 \\
CL-Proxy (Eq.~\ref{eq:clproxy}) & 3 & \textbf{0.90} & \textbf{3} \\
\bottomrule
\end{tabular}
\end{table}

CL-Proxy matches the full PDMS exactly in both Spearman $\rho$ and number of pairwise ranking inversions, despite using only 60\% of the sub-metrics and a substantially simpler functional form. On this dataset, \textbf{TTC and Comfort add little marginal information beyond NC, DAC, and EP for closed-loop ranking}. We do not claim TTC and Comfort are intrinsically uninformative: rather, among the current generation of top methods (Table~\ref{tab:submetrics}), TTC ranges narrowly from 92.9 to 97.2 and Comfort is essentially saturated ($\geq{}99.9\%$), so neither has enough variance to discriminate methods. As future planners diverge in collision-avoidance margin or comfort behavior, the marginal value of TTC and C may grow, and CL-Proxy's equivalence to PDMS should be re-tested. EP alone, by contrast, falls short of PDMS even on our data ($\rho{=}0.83$ vs.\ $0.90$): the NC and DAC penalty gates remain useful for separating safety-violating outliers.

We also observed that no simple monotonic combination of the four sub-metrics (NC, DAC, TTC, EP)---whether linear, multiplicative, or power-law---exceeds $\rho{=}0.90$ on our paired data. The bottleneck is structural: SafeDrive simultaneously achieves the top rank in NC, DAC, and TTC yet ranks only 5th in DS, an inversion that no aggregation of these sub-metrics can repair. CL-Proxy is therefore not just a simplification but, on this dataset, an \emph{optimal} simple aggregator within the family of monotonic sub-metric combinations.

From a practical standpoint, CL-Proxy retains PDMS's interpretable gate structure and can serve as a transparent, lightweight diagnostic alongside the official PDMS---particularly useful when sub-metric breakdowns are reported but the full PDMS is not.

\subsection{Proposed Guidelines for Open-Loop Benchmark Design}

Based on our analysis, we propose the following guidelines:

\begin{enumerate}[leftmargin=*]
  \item \textbf{Report sub-metrics, not just aggregates.} The aggregate PDMS/EPDMS hides critical information about the safety--progress trade-off. Leaderboards should prominently display individual sub-metrics.
  \item \textbf{Prefer simpler aggregations when possible.} Our CL-Proxy result shows that on $n{=}8$ paired methods, 3 of the 5 PDMS sub-metrics suffice to recover the full aggregate's rank correlation with closed-loop DS. TTC and Comfort exhibit narrow inter-method variance among current top planners (TTC spans 4.3 points, Comfort is saturated near 100\%) and therefore add little discriminative value in the present sample. This argues for distinguishing \emph{diagnostic} sub-metrics (kept for fault analysis) from \emph{ranking} sub-metrics, and for periodically re-validating which is which as the leaderboard evolves.
  \item \textbf{Introduce a slow-driving penalty.} Unlike Bench2Drive, which penalizes overly conservative driving via the ``Too Slow'' infraction, NAVSIM lacks an equivalent penalty. Adding such a penalty would better align open-loop evaluation with closed-loop outcomes, addressing the safety--progress trade-off at the source.
  \item \textbf{Calibrate with paired data.} As more methods report both benchmarks, periodic re-calibration of the open-to-closed-loop mapping becomes possible and should be standard practice.
\end{enumerate}

\section{Discussion}
\label{sec:discussion}

\paragraph{The Snowball Effect: Why Small Open-Loop Deviations Compound.}
A candidate mechanism for the residual open-to-closed-loop gap is what we term the \emph{snowball effect}: small trajectory deviations that are inconsequential in NAVSIM's 4-second open-loop evaluation can cascade into significant failures over multi-minute closed-loop routes. For example, a planner that drives ``slightly slow'' for 4~seconds may receive only a minor EP penalty in NAVSIM ($\sim$2\% reduction). However, in Bench2Drive's closed-loop evaluation, the same tendency compounds: the vehicle misses a green light phase $\rightarrow$ waits through a full red cycle $\rightarrow$ falls behind schedule $\rightarrow$ triggers ``Too Slow'' penalties ($p = 0.70$) $\rightarrow$ eventually times out ($p = 0.70$), resulting in a DS near zero for that route. This exponential amplification of small biases is a plausible reason why EP---which directly measures progress tendency---tracks closed-loop DS more closely than safety metrics like NC, which capture only instantaneous collision avoidance.

The snowball effect is also consistent with the PDMS--DS correlation, while strong ($\rho = 0.90$), not being perfect: NAVSIM's single-step evaluation cannot capture the temporal compounding that dominates closed-loop outcomes.

\paragraph{The Reactivity Gap.}
NAVSIM evaluates trajectories against \emph{recorded} agent behaviors: other vehicles follow their logged trajectories regardless of what the ego vehicle does. In contrast, Bench2Drive simulates \emph{reactive} agents that adjust their behavior in response to the ego's actions. This asymmetry has two consequences:

\begin{itemize}[leftmargin=*]
  \item \textbf{Aggressive maneuvers are under-penalized in open-loop.} A lane change that narrowly cuts in front of another vehicle may appear safe in NAVSIM (the other vehicle maintains its recorded trajectory), but in Bench2Drive the cut-off vehicle may brake suddenly or swerve, creating new collision scenarios.
  \item \textbf{Defensive driving is over-rewarded in open-loop.} Methods like SafeDrive that avoid all proximity to other agents receive high NC and TTC scores in NAVSIM, but in closed-loop the same conservatism prevents the ego from navigating through gaps in traffic, leading to deadlocks.
\end{itemize}

This reactivity gap is structurally irreducible within any open-loop framework---it can only be mitigated, not eliminated, by better metric design.

\paragraph{Confounders in Cross-Architecture Comparisons.}
An important caveat is that the 8 paired methods differ not only in their planning algorithms but also in backbone architectures (ResNet-34 vs.\ ResNet-50 vs.\ ViT-L), sensor modalities (camera-only vs.\ camera+LiDAR), and training data configurations. As Waymo~\cite{waymo2025scaling} demonstrated, within a single architecture family, open-loop loss and closed-loop failure rates follow a power-law relationship---but this relationship breaks down across architectures. Our $\rho{=}0.90$ is therefore a cross-architecture correlation, not an estimate of the within-architecture metric--DS relationship; the two need not coincide and could differ in either direction. Methods like RAP~\citep{rap2025} and LEAD~\citep{lead2025}, which report both benchmarks but with different model configurations (e.g., different backbones or sensor setups for NAVSIM vs.\ Bench2Drive), were excluded from our paired dataset to avoid mixing such configuration drift into the correlation.

\paragraph{Limitations.}
Our analysis has several limitations:

\begin{enumerate}[leftmargin=*]
  \item \textbf{Sample size.} The paired dataset contains 8 methods with complete PDMS--DS data and sub-metric breakdowns. While the correlations are statistically significant ($p < 0.01$), larger paired datasets would strengthen confidence. We expect this dataset to grow as more methods adopt both benchmarks.
  \item \textbf{Bench2Drive variance.} Closed-loop evaluation on CARLA exhibits run-to-run variance ($\sim$5 DS for the same model with different seeds). The DS values we use are from published (typically single-run) evaluations, potentially introducing noise.
  \item \textbf{NAVSIM v1 vs.\ v2.} Some methods report PDMS (v1) while others report EPDMS (v2). We use PDMS where available as it is reported by more methods, but the transition to EPDMS introduces systematic differences in sub-metric composition.
  \item \textbf{No causal analysis.} Our study is purely correlational. We do not claim that improving EP \emph{causes} higher DS; rather, both may reflect an underlying capability (e.g., accurate speed planning) that manifests in both benchmarks.
\end{enumerate}

\paragraph{Connection to Waymo Scaling Laws.}
The Waymo team~\cite{waymo2025scaling} demonstrated that within a single architecture family, open-loop loss and closed-loop failures follow a power law: $\eta(C) = aC^b + K_\eta$. Our finding is complementary: we show that \emph{across} architectures, the correlation exists at the sub-metric level (especially EP) but is obscured when compressed into a single aggregate score. The combination of these findings suggests a two-level approach: use scaling laws within an architecture, and sub-metric regression across architectures.

\paragraph{Broader Impact.}
Reliable open-loop evaluation of autonomous driving systems could meaningfully accelerate development cycles. However, over-reliance on any proxy metric carries risks: if the community optimizes for an imperfect open-loop proxy, it may develop systems that ``game'' the proxy without genuine closed-loop improvement. Our sub-metric-level analysis provides a more transparent framework that is harder to game than a single scalar score.

\section{Conclusion}
\label{sec:conclusion}

We present the first systematic cross-benchmark correlation study between NAVSIM open-loop metrics and Bench2Drive closed-loop performance, based on 8 paired methods. Our key findings are:

\begin{enumerate}[leftmargin=*]
  \item Traditional open-loop metrics (ADE/FDE) remain uncorrelated with closed-loop driving quality ($\rho = -0.36$, $p = 0.43$), confirming prior work with newer methods.
  \item NAVSIM's safety-aware PDMS shows strong positive correlation with Bench2Drive DS ($\rho = 0.90$, $p = 0.002$), but with non-monotonic ranking inversions driven by the safety--progress trade-off.
  \item Among individual sub-metrics, Ego Progress (EP) is the strongest single predictor of closed-loop DS ($\rho = 0.83$), while the collision metric NC shows only weak-to-moderate correlation ($\rho = 0.45$) due to ceiling effects despite extreme safety optimization.
  \item The \emph{snowball effect}---where small open-loop deviations compound into closed-loop failures via missed traffic signals, timeouts, and blocking---is a candidate mechanism consistent with EP being a stronger predictor than the safety-only sub-metrics (NC, DAC, TTC), although our correlational analysis does not establish this causally.
  \item A simplified CL-Proxy score using only NC, DAC, and EP matches the rank correlation of the full 5-sub-metric PDMS ($\rho = 0.90$, 3 inversions) on our $n{=}8$ paired data, suggesting that TTC and Comfort add little marginal ranking information among current top methods---a finding that should be re-validated as the leaderboard diversifies.
\end{enumerate}

These findings argue strongly for \emph{sub-metric-level analysis} in open-loop evaluation, and for explicit calibration between open-loop and closed-loop benchmarks. As the paired dataset grows with future publications, we expect the open-to-closed-loop mapping to become increasingly reliable.


{\small
\bibliography{Bib}
}

\end{document}